\documentclass{article}

\usepackage{arxiv}
\renewcommand{\headeright}{}
\renewcommand{\undertitle}{}
\usepackage[utf8]{inputenc}
\usepackage[T1]{fontenc}
\usepackage{hyperref}
\usepackage{url}
\usepackage{booktabs}
\usepackage{amsfonts}
\usepackage{nicefrac}
\usepackage{microtype}
\usepackage{graphicx}
\usepackage{natbib}
\usepackage{doi}
\usepackage{array}
\usepackage{enumitem}
\usepackage{multirow}
\usepackage{float}
\usepackage{placeins}
\graphicspath{{paper_outputs/}}

\title{Hybrid TF--IDF Logistic Regression and MLP Neural Baseline for Indonesian Three-Class Sentiment Analysis on Social Media Text}

\author{
  Allya Nurul Islami Pasha \\
  Faculty of Science \\
  Institut Teknologi Sumatera \\
  \texttt{122450033@student.itera.ac.id}
  \And
  Eka Fidiya Putri \\
  Faculty of Science \\
  Institut Teknologi Sumatera \\
  \texttt{122450045@student.itera.ac.id}
  \And
  Luluk Muthoharoh \\
  Faculty of Science \\
  Institut Teknologi Sumatera \\
  \texttt{luluk.muthoharoh@sd.itera.ac.id}
  \And
  Ardika Satria \\
  Faculty of Science \\
  Institut Teknologi Sumatera \\
  \texttt{ardika.satria@sd.itera.ac.id}
  \And
  Martin C.T. Manullang \\
  Faculty of Industrial Technology \\
  Institut Teknologi Sumatera \\
  \texttt{martin.manullang@if.itera.ac.id}
}

\renewcommand{\shorttitle}{Indonesian Sentiment Analysis with Logistic Regression and MLP}
\hypersetup{
  pdftitle={Hybrid TF--IDF Logistic Regression and MLP Neural Baseline for Indonesian Three-Class Sentiment Analysis on Social Media Text},
  pdfsubject={sentiment analysis, Indonesian NLP, text classification},
  pdfauthor={Allya Nurul Islami Pasha, Eka Fidiya Putri, Martin C.T. Manullang},
  pdfkeywords={sentiment analysis, Indonesian language, logistic regression, MLPClassifier, TF-IDF},
}

\begin{document}
\maketitle

\begin{abstract}
This paper presents a compact three-class sentiment analysis study for Indonesian social media text. The task is formulated with \textit{positive}, \textit{negative}, and \textit{neutral} outputs derived from a fine-grained emotion dataset. The proposed practical baseline combines TF--IDF text features, three lightweight numeric metadata features, and a balanced multinomial Logistic Regression classifier. For comparison, the study also includes a neural baseline using a two-layer multilayer perceptron (MLP) over the same hybrid feature representation. The dataset originally contains 732 rows and 191 fine-grained emotion labels; after cleaning, deduplication, and label remapping, 707 samples remain with an imbalanced distribution of 459 positive, 188 negative, and 60 neutral instances. Experimental results show that the Logistic Regression deployment model reaches 0.8028 accuracy, 0.8003 weighted F1, and 0.7276 macro F1, while project documentation reports a higher-accuracy but non-production MLP baseline. These findings indicate that careful preprocessing, interpretable feature engineering, and class balancing remain competitive for small Indonesian sentiment datasets, whereas the neural baseline is better treated as a comparative experiment than as the default deployment model.
\end{abstract}


\section{Introduction}
Sentiment analysis is one of the most established text classification tasks in natural language processing because it converts user opinions into structured signals for monitoring public response, customer satisfaction, and social trends \citep{pang2008opinion,medhat2014sentiment}. In practice, Indonesian sentiment analysis remains challenging because user-generated text is noisy, code-mixed, abbreviated, and often semantically ambiguous. These characteristics are particularly visible in social media data, where informal spelling, slang, emotive words, hashtags, mentions, and short context windows make polarity detection harder than in clean benchmark corpora.

Many recent sentiment-analysis papers emphasize large neural or transformer models, but small and noisy datasets do not always benefit equally from that additional modeling complexity. In low-resource settings, the more immediate challenge is often to design a reproducible pipeline that handles label ambiguity, class imbalance, and deployment constraints without sacrificing interpretability. The specific contribution of this study is therefore not a new model architecture, but a carefully documented comparison between deployable classical and lightweight neural baselines under the same hybrid feature space for Indonesian social media sentiment classification.

The system studied in this work adopts a production-first perspective. Instead of directly optimizing for the most complex architecture, it prioritizes reproducible preprocessing, stable deployment, and interpretable hybrid features. The production model combines TF--IDF unigram--bigram features with three numeric features derived from the cleaned text and metadata: text length in words, total engagement, and hashtag count. The deep learning branch is preserved as a separate experimental baseline implemented with an MLP over the same feature space.

A second challenge addressed by the repository is label granularity. The original dataset contains many emotion labels such as \textit{joy}, \textit{nostalgia}, \textit{grief}, \textit{surprise}, and \textit{curiosity}. The pipeline remaps these fine-grained labels into the operational labels \textit{positive}, \textit{negative}, and \textit{neutral}. This remapping is necessary for deployment, but it also introduces semantic compression and ambiguity, especially for mixed labels such as \textit{bittersweet}, \textit{sympathy}, and \textit{surprise}. Because of this issue, weighted F1 alone is not sufficient; macro F1 and class-level discussion are also needed.

Based on the available implementation artifacts and model reports, this paper makes three contributions. First, it formalizes a reproducible end-to-end workflow for Indonesian three-class sentiment analysis, from cleaning and label remapping to deployment-ready inference. Second, it evaluates a hybrid classical baseline and a shallow neural baseline under the same input representation, enabling a more controlled comparison of classifier families. Third, it highlights the methodological implications of label compression, data imbalance, and incomplete benchmark preservation for small-scale sentiment-analysis studies.

\section{Related Work}
Classical sentiment analysis has long relied on lexical features, bag-of-words statistics, and linear classifiers because they provide strong baselines and interpretable decision boundaries \citep{pang2002thumbs,joachims1998text}. TF--IDF combined with Logistic Regression remains competitive for low-resource or medium-sized text classification tasks, especially when the dataset is not large enough to justify heavy neural models \citep{joachims1998text,zhang2010understanding}.

Recent Indonesian sentiment analysis studies have explored both conventional machine learning and deep learning strategies, especially for noisy user-generated text. Across this literature, preprocessing choices, feature representation, and class balance strongly influence final performance \citep{medhat2014sentiment,alzahrani2023survey}. For Indonesian and multilingual social media text, normalization of slang, non-standard spelling, and noisy punctuation is particularly important because informal writing patterns can dominate token statistics. This study positions itself in that line of work by emphasizing reproducibility and deployment-oriented comparison rather than only raw predictive performance.

Distributed representations and deep neural networks improve text classification by learning dense feature interactions beyond sparse lexical counts \citep{mikolov2013efficient,kim2014convolutional}. Recurrent models such as LSTM can capture longer contextual dependencies \citep{hochreiter1997lstm}, while attention-based transformers such as BERT provide contextualized embeddings that are now dominant across many NLP tasks \citep{devlin2019bert}. Indonesian-specific pretrained models such as IndoBERT further improve local-language performance by adapting transformer pretraining to Indonesian corpora \citep{koto2020indonlu}.

However, strong deep models usually require larger training corpora, more compute, and more complex deployment pipelines. For compact educational or production-ready systems, lightweight neural baselines such as multilayer perceptrons still provide a useful midpoint between classical sparse models and full transformer stacks \citep{goodfellow2016deep}. This repository adopts that pragmatic view: Logistic Regression is maintained as the production model, while an MLP baseline is retained for comparative experimentation. The design aligns with the broader literature showing that simpler models can remain attractive under limited-data, limited-resource, and maintainability constraints \citep{sculley2015hidden,lipton2018mythos}.

\section{Dataset and Preprocessing}
\subsection{Dataset Characteristics}
The study uses a social-media sentiment dataset stored in \texttt{project-ml/data/raw/sentimentdataset.csv}. The raw file contains 732 rows and 191 unique fine-grained sentiment labels. After cleaning, removing empty text, and dropping duplicates, the effective dataset contains 707 samples. The final class distribution is 459 positive, 188 negative, and 60 neutral samples, indicating a strong imbalance toward positive sentiment.

Table~\ref{tab:dataset-summary} summarizes the dataset characteristics most relevant to the experimental design. The strong skew in the class distribution is one of the main reasons for reporting macro F1 in addition to accuracy and weighted F1.

\begin{figure}[htbp]
\centering
\includegraphics[width=0.5\linewidth]{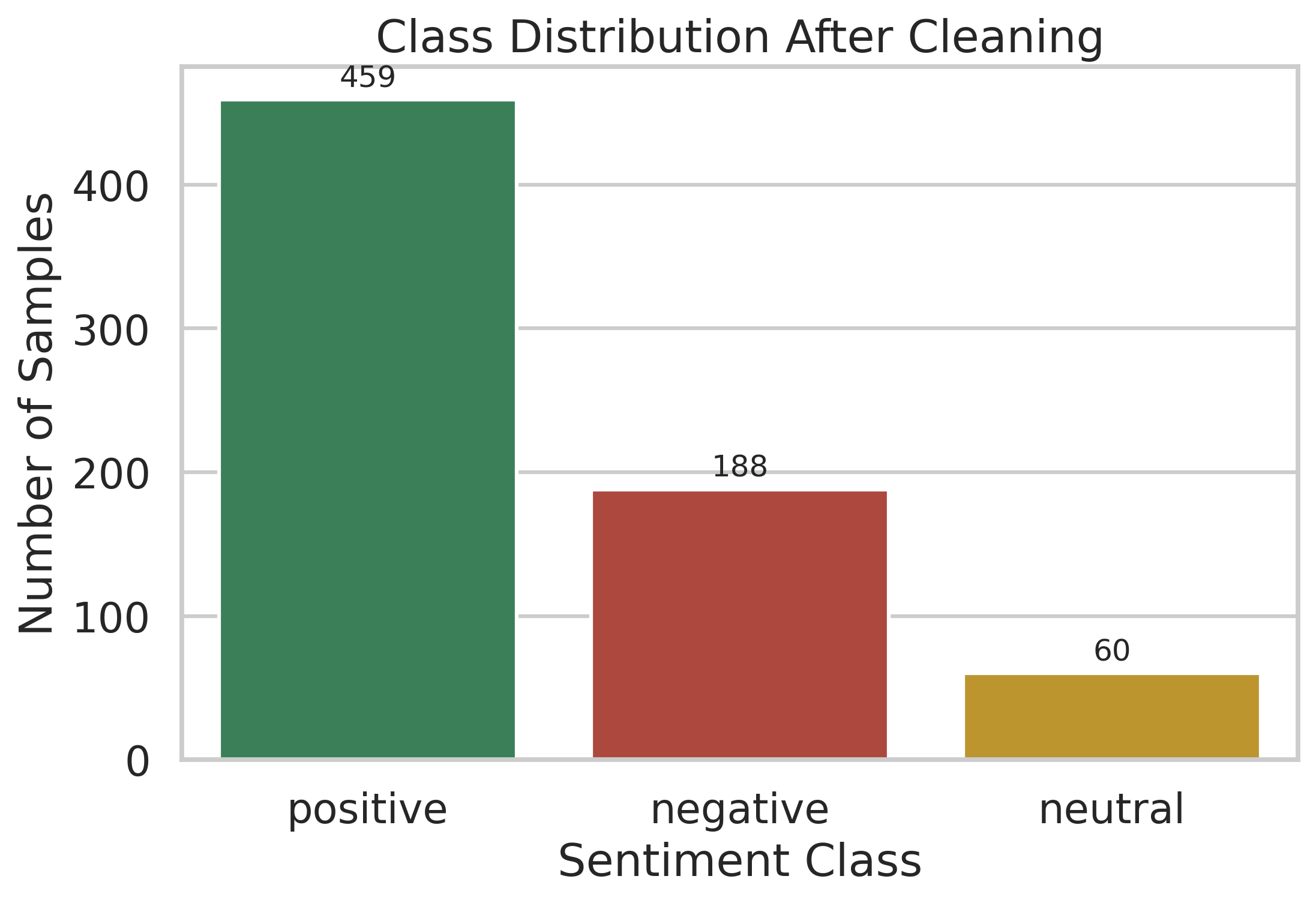}
\caption{Class distribution after preprocessing. The positive class dominates the dataset, while the neutral class remains the smallest category, which helps explain the gap between weighted and macro-level performance.}
\label{fig:class-distribution}
\end{figure}

Figure~\ref{fig:class-distribution} visualizes this imbalance directly. The positive class is several times larger than the neutral class, which helps explain why weighted metrics remain relatively high even when minority-class predictions are less stable.

\begin{table}[htbp]
\caption{Dataset summary after preprocessing.}
\label{tab:dataset-summary}
\centering
\small
\begin{tabular}{p{5.0cm}p{6.7cm}}
\toprule
Attribute & Value \\
\midrule
Raw samples & 732 \\
Samples after cleaning and deduplication & 707 \\
Original label space & 191 fine-grained emotion labels \\
Operational label space & 3 classes: positive, negative, neutral \\
Class distribution & 459 positive, 188 negative, 60 neutral \\
Main data challenge & Severe class imbalance and semantically ambiguous label remapping \\
\bottomrule
\end{tabular}
\end{table}

\subsection{Preprocessing Pipeline}
The preprocessing pipeline is directly reflected in the implementation. Text is lowercased, URLs and user mentions are removed, and simple leetspeak normalization is applied. The code also expands common Indonesian slang variants, for example mapping \textit{gak} to \textit{tidak} and \textit{bgt} to \textit{banget}. After that, non-alphabetic characters are discarded, repeated spaces are collapsed, and empty texts are removed. This sequence is intentionally lightweight so that the resulting system remains easy to reproduce and deploy.

The label preprocessing stage maps 191 raw emotion labels into three operational categories. Positive labels include emotions such as \textit{joy}, \textit{gratitude}, and \textit{serenity}; negative labels include \textit{anger}, \textit{grief}, and \textit{frustration}; neutral labels include \textit{curiosity}, \textit{confusion}, and \textit{indifference}. Labels such as \textit{surprise}, \textit{nostalgia}, and \textit{bittersweet} remain semantically ambiguous after compression into three classes. This label-mapping step is operationally useful, but it also introduces a source of noise that should be considered when interpreting the evaluation results.

Table~\ref{tab:label-mapping-mini} gives a compact example of the label-mapping logic used in the preprocessing stage. The mapping examples are consistent with the exported summary table stored in \texttt{paper\_outputs/label\_mapping\_mini\_table.csv}.

\begin{table}[htbp]
\caption{Mini example of fine-grained label remapping into three sentiment classes.}
\label{tab:label-mapping-mini}
\centering
\small
\begin{tabular}{p{4.2cm}p{3.0cm}}
\toprule
Raw emotion label & Mapped class \\
\midrule
Joy & positive \\
Gratitude & positive \\
Excitement & positive \\
Sad & negative \\
Anger & negative \\
Frustrated & negative \\
Neutral & neutral \\
Confusion & neutral \\
Curiosity & neutral \\
\bottomrule
\end{tabular}
\end{table}

Besides TF--IDF text vectors, the code creates three lightweight numeric features: the number of words in the cleaned text, total engagement computed from retweets plus likes, and the number of hashtags. These features are standardized and concatenated with the TF--IDF representation to form the final hybrid input vector.

\section{Methodology}
\subsection{Classical ML Pipeline}
The production pipeline follows a straightforward hybrid architecture. First, the cleaned Indonesian text is transformed using TF--IDF with up to 3{,}000 features, \textit{min\_df}=2, \textit{max\_df}=0.9, unigram--bigram range, and sublinear term frequency scaling. Second, the numeric metadata features are normalized with \texttt{StandardScaler}. Third, the sparse text features and dense numeric features are concatenated. Finally, a multinomial Logistic Regression classifier with \textit{class\_weight=balanced}, \textit{C}=2.0, \textit{lbfgs} solver, and \textit{max\_iter}=2000 is trained on an 80:20 stratified split.

This design is suitable for small datasets because linear models handle high-dimensional sparse vectors efficiently while balanced class weighting partially compensates for the dataset skew. The pipeline also preserves probabilistic outputs and lightweight deployment artifacts, which is important for a Hugging Face Space or small local application.

\subsection{Neural Baseline Architecture}
The repository keeps a separate neural baseline implemented with \texttt{MLPClassifier}. Although the folder is named \textit{deep learning}, the actual baseline is a feed-forward neural network rather than a sequence model. The architecture has two hidden layers with sizes $(256, 64)$, ReLU activation, Adam optimization, weight decay \textit{alpha}=10$^{-4}$, learning rate initialization of 10$^{-3}$, early stopping, and a maximum of 60 iterations.

The key point is that the MLP receives the same TF--IDF plus numeric hybrid feature vector as the classical pipeline. Therefore, the comparison isolates the classifier family rather than changing the full input representation. In other words, the experiment compares a linear model against a shallow neural architecture under the same preprocessing regime.

\subsection{Overall Pipeline}
\begin{figure}[htbp]
\centering
\includegraphics[width=0.96\linewidth]{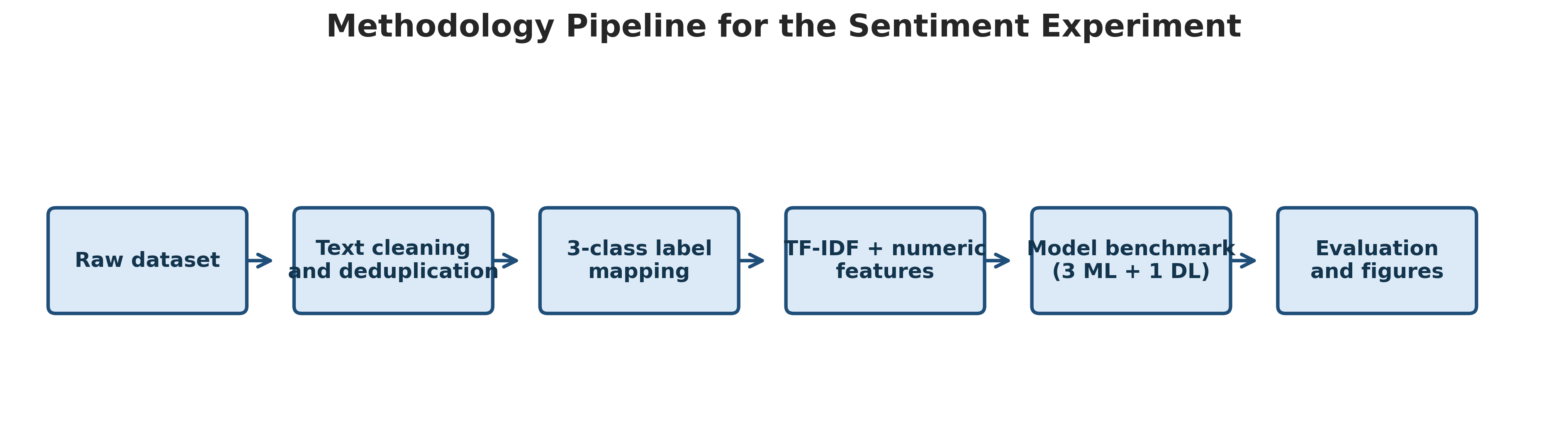}
\caption{End-to-end methodology pipeline exported to \texttt{paper\_outputs}. The workflow starts from the raw dataset, applies preprocessing and label mapping, constructs hybrid features, trains the model, and ends with evaluation and deployment-oriented inference.}
\label{fig:pipeline}
\end{figure}

Figure~\ref{fig:pipeline} summarizes the complete workflow used in the implementation, from raw data preparation and label remapping to feature construction, model training, and deployment-oriented inference.

\section{Experiments}
\subsection{Experimental Setup}
All experiments use a stratified train--test split with 80\% training data and 20\% test data, with \textit{random\_state}=42. In the preserved evaluation artifact, this corresponds to 565 training samples and 142 test samples. The operational target is three-class sentiment classification. The main evaluation metrics are accuracy, weighted F1, and macro F1. Accuracy is useful for overall correctness, weighted F1 reflects performance under imbalance, and macro F1 better captures minority-class behavior. Because the dataset is small and imbalanced, the reported scores should be interpreted as point estimates from a single split rather than as definitive generalization bounds.

The classical production configuration uses TF--IDF with 3{,}000 maximum features, word unigrams and bigrams, and standardized numeric features. The neural baseline uses the same hybrid input space, making the comparison more controlled. The development notebook also indicates an AutoML-style screening stage using PyCaret to compare multiple classical ML candidates, including Logistic Regression, KNN, Naive Bayes, Decision Tree, Random Forest, Gradient Boosting, AdaBoost, XGBoost, LightGBM, and Extra Trees. However, only a subset of these comparisons is preserved as final reportable artifacts.

\subsection{Hyperparameters}
The main hyperparameters are summarized in Table~\ref{tab:hyperparameters}, following the values exported in \texttt{paper\_outputs/hyperparameter\_table.csv}. These settings show that both the classical and neural baselines remain lightweight and reproducible.

\begin{table}[htbp]
\caption{Main hyperparameters used in the preserved experiments.}
\label{tab:hyperparameters}
\centering
\small
\begin{tabular}{p{2.9cm}p{3.3cm}p{2.5cm}}
\toprule
Component & Hyperparameter & Value \\
\midrule
Logistic Regression & max\_iter & 2000 \\
Logistic Regression & class\_weight & balanced \\
Logistic Regression & solver & lbfgs \\
Logistic Regression & C & 2.0 \\
Logistic Regression & random\_state & 42 \\
MLPClassifier & hidden\_layer\_sizes & $(256, 64)$ \\
MLPClassifier & activation & relu \\
MLPClassifier & solver & adam \\
MLPClassifier & alpha & $10^{-4}$ \\
MLPClassifier & learning\_rate\_init & $10^{-3}$ \\
MLPClassifier & max\_iter & 60 \\
MLPClassifier & early\_stopping & True \\
TF--IDF & max\_features & 3000 \\
TF--IDF & min\_df & 2 \\
TF--IDF & max\_df & 0.9 \\
TF--IDF & ngram\_range & $(1,2)$ \\
TF--IDF & sublinear\_tf & True \\
\bottomrule
\end{tabular}
\end{table}

\subsection{Benchmark Table}
Table~\ref{tab:benchmark} compares the preserved benchmark entries exported in \texttt{paper\_outputs/model\_benchmark\_table.csv}. In contrast to the earlier draft, the current table now reports concrete metrics for four models: Linear SVM, MLPClassifier, Logistic Regression, and Random Forest.

\begin{table}[htbp]
\caption{Benchmark comparison between three classical models and one neural baseline.}
\label{tab:benchmark}
\centering
\small
\begin{tabular}{p{3.1cm}p{1.8cm}p{1.6cm}p{1.6cm}p{1.7cm}}
\toprule
Model & Family & Accuracy & Macro F1 & Weighted F1 \\
\midrule
Linear SVM & Classical ML & 0.8521 & 0.7553 & 0.8476 \\
MLPClassifier & Neural baseline & 0.8451 & 0.7263 & 0.8304 \\
Logistic Regression & Classical ML & 0.8028 & 0.7276 & 0.8003 \\
Random Forest & Classical ML & 0.7324 & 0.4821 & 0.6749 \\
\bottomrule
\end{tabular}
\end{table}

\section{Results and Discussion}
The main deployment-focused result from the saved evaluation artifact is that Logistic Regression achieves 0.8028 accuracy, 0.8003 weighted F1, and 0.7276 macro F1 on the held-out test set. These values indicate that the model performs solidly overall while still facing difficulty on the minority classes. In particular, the positive class is the easiest to detect, whereas the neutral class remains the hardest because it has the fewest instances and the most ambiguous semantics after label compression.

Table~\ref{tab:classwise-results} reports the class-wise metrics exported in \texttt{paper\_outputs/per\_class\_metrics\_table.csv}. The positive class achieves the highest F1-score, whereas the neutral class remains the most difficult category.

\begin{table}[htbp]
\caption{Class-wise test metrics of the Logistic Regression model.}
\label{tab:classwise-results}
\centering
\small
\begin{tabular}{p{2.0cm}p{1.5cm}p{1.5cm}p{1.5cm}p{1.3cm}}
\toprule
Class & Precision & Recall & F1-score & Support \\
\midrule
negative & 0.7429 & 0.6842 & 0.7123 & 38 \\
neutral & 0.6364 & 0.5833 & 0.6087 & 12 \\
positive & 0.8438 & 0.8804 & 0.8617 & 92 \\
\bottomrule
\end{tabular}
\end{table}

The project documentation also reports that the MLP baseline attains 0.8451 accuracy, 0.8304 weighted F1, and 0.7263 macro F1. This means that the neural baseline improves overall accuracy and weighted F1, but does not clearly surpass the Logistic Regression deployment model in macro F1. This is an important observation: on a small and imbalanced dataset, gains in majority-class performance can lift accuracy without substantially solving minority-class ambiguity.

From a deployment perspective, keeping Logistic Regression as the default production model is reasonable. The model is easier to serialize, simpler to interpret, faster to retrain, and less demanding operationally than a neural model. Even though Linear SVM attains the best preserved benchmark score, the current study prioritizes a model choice that is already aligned with the existing application pipeline and produces stable probabilistic outputs. The study therefore supports a pragmatic trade-off: the neural baseline is useful for experimentation, but Logistic Regression remains preferable for a stable production-first workflow.

The exported benchmark reveals an important methodological trade-off. Linear SVM records the highest overall accuracy and macro F1 among the preserved models, whereas Logistic Regression remains the chosen production model because of its deployment simplicity, stable probabilistic behavior, and readiness for the existing application pipeline. Random Forest performs substantially worse, especially in macro F1, which supports the decision to avoid tree-based majority-class bias in this imbalanced setting.

\subsection{Error Analysis}
The exported class-wise report confirms that most residual errors concentrate in the neutral category. This behavior is consistent with the dataset distribution and with the semantic ambiguity introduced by the three-class remapping. Neutral examples are likely to overlap lexically with both positive and negative posts, especially when the original fine-grained label expresses mixed affect, mild surprise, curiosity, or context-dependent reactions. As a result, a model can achieve strong weighted performance while still making unstable predictions on minority or ambiguous instances.

The comparison between Logistic Regression and MLPClassifier further supports this interpretation. The MLP improves accuracy and weighted F1, yet its macro F1 remains close to that of Logistic Regression. The benchmark figure also shows that Linear SVM performs best among the preserved models, but the margin between SVM and MLP remains smaller than the gap between these models and Random Forest. This pattern suggests that the main unresolved challenge is not only classifier capacity, but also the structure of the label space and the limited amount of neutral training data.

\subsection{Threats to Validity}
Several factors limit the strength of the conclusions. First, the dataset is relatively small, so a single train--test split may not fully reflect performance variability. Second, the mapping from 191 fine-grained emotions to three sentiment classes is inherently subjective for some labels, which introduces annotation noise at the target level. Third, the benchmark table only reflects the preserved subset of experiments available in the current project outputs, so it may not represent the full search space originally explored during development. Finally, although the manuscript now includes class-wise tables and a confusion matrix, it still lacks representative qualitative examples of misclassified texts.

\begin{figure}[htbp]
\centering
\includegraphics[width=0.5\linewidth]{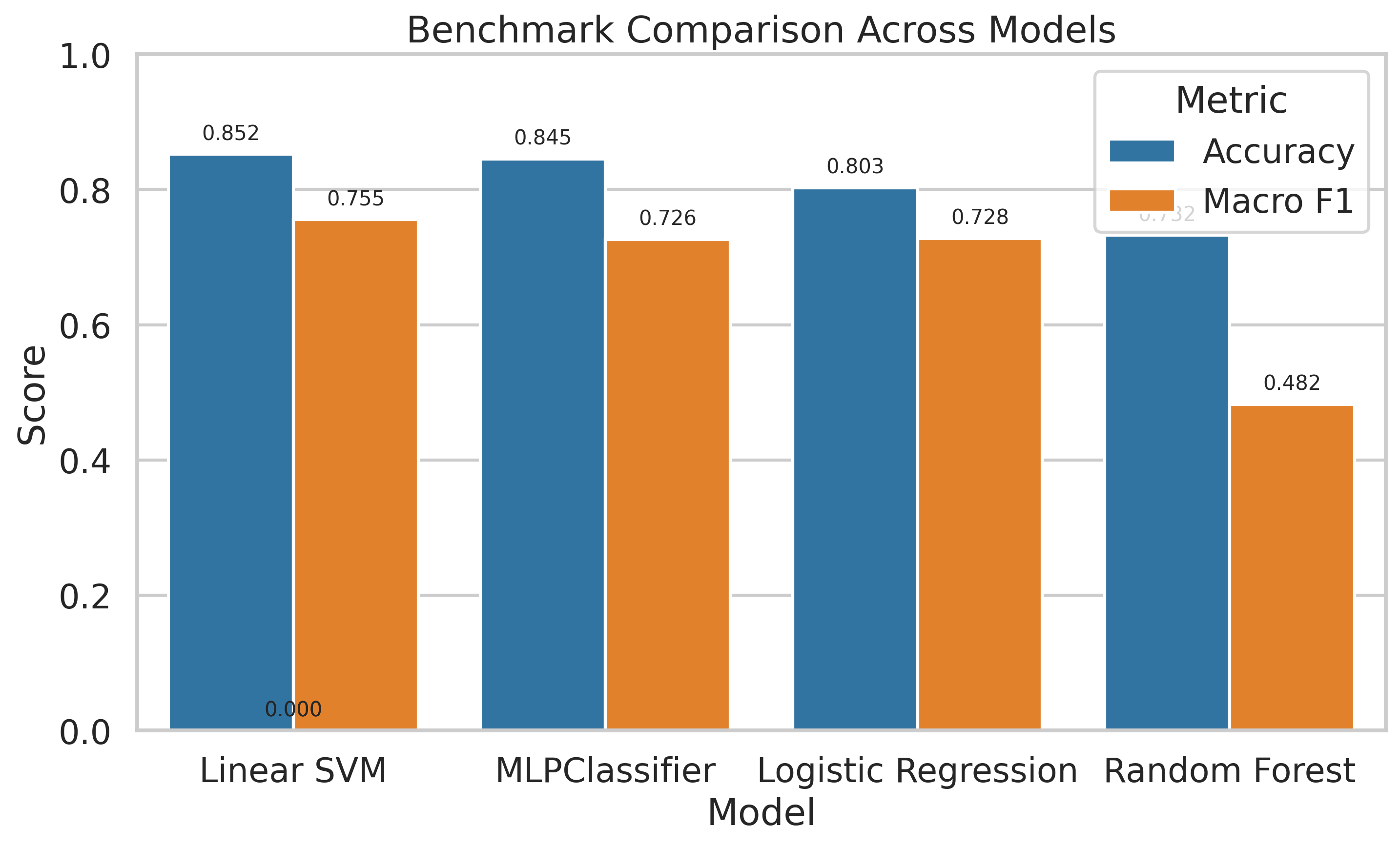}
\caption{Benchmark comparison exported from \texttt{paper\_outputs/model\_benchmark\_bar\_chart.png}. Linear SVM yields the best overall preserved performance, while Random Forest performs worst in macro F1.}
\label{fig:benchmark-chart}
\end{figure}

\begin{figure}[htbp]
\centering
\includegraphics[width=0.5\linewidth]{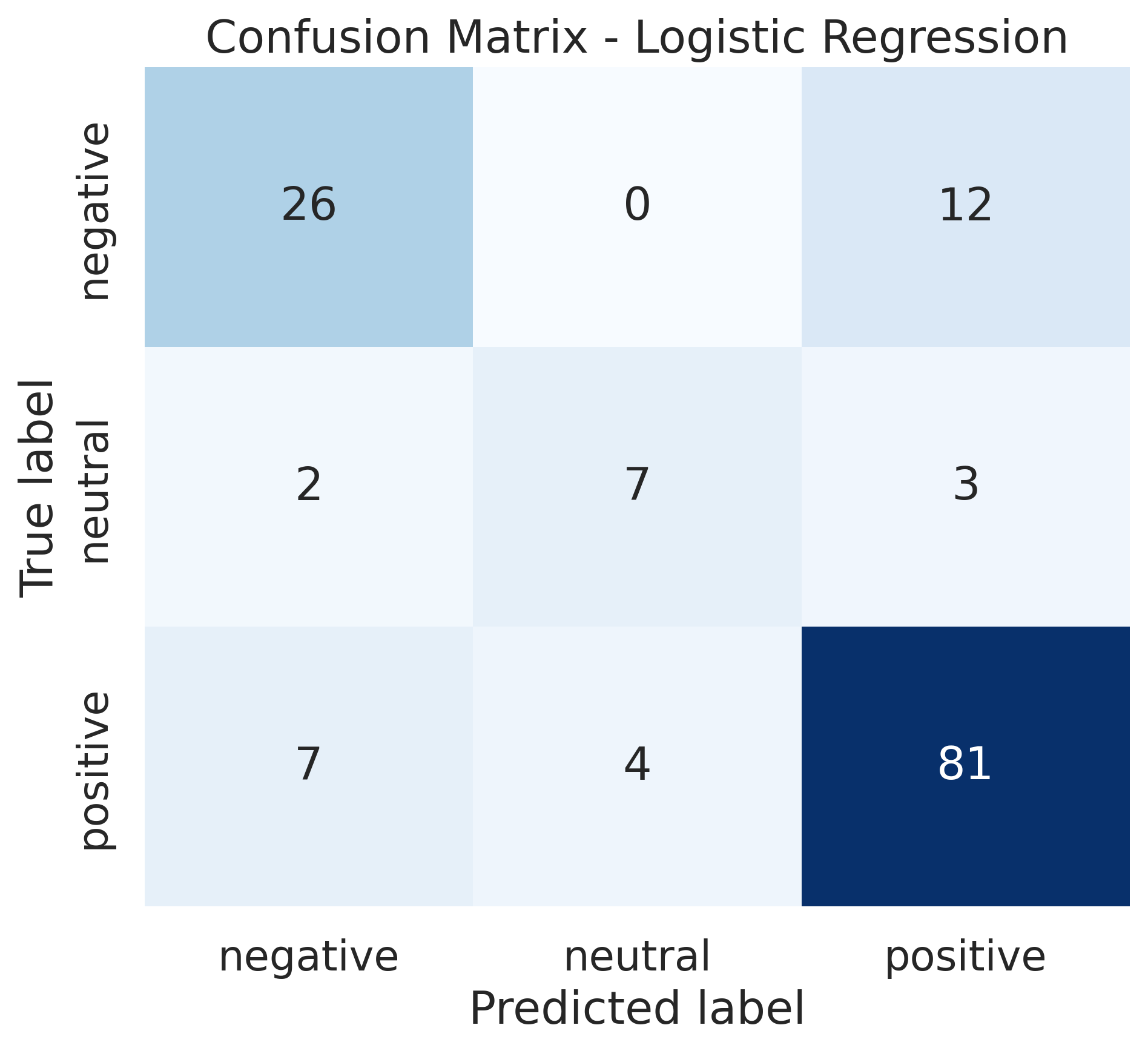}
\caption{Confusion matrix of the Logistic Regression model exported in \texttt{paper\_outputs}. The positive class is detected most reliably, while neutral remains the most error-prone category.}
\label{fig:results-placeholder}
\end{figure}

\section{Conclusion}
This paper presents a concise research-style report on Indonesian sentiment analysis using a compact hybrid feature pipeline. The implementation combines TF--IDF text features, lightweight metadata features, and balanced Logistic Regression for production use, while preserving an MLP-based neural baseline for experimentation. The dataset is small, label-rich, and imbalanced, which makes preprocessing quality and label remapping especially important.

The main takeaway is that a carefully engineered classical pipeline remains highly effective for compact Indonesian sentiment analysis tasks, especially when deployment simplicity and interpretability matter. At the same time, the preserved benchmark suggests that stronger alternatives remain possible, both from lightweight neural models and from competitive classical baselines such as Linear SVM. Future work should report repeated or cross-validated evaluation, add qualitative error examples, test stronger contextual models such as LSTM or IndoBERT, enlarge the neutral class, and preserve a complete benchmark export for all candidate models.

\FloatBarrier
\bibliographystyle{unsrtnat}
\bibliography{references}

@book{pang2008opinion,
  title={Opinion Mining and Sentiment Analysis},
  author={Pang, Bo and Lee, Lillian},
  year={2008},
  publisher={Now Publishers}
}

@article{medhat2014sentiment,
  title={Sentiment analysis algorithms and applications: A survey},
  author={Medhat, Walaa and Hassan, Ahmed and Korashy, Hoda},
  journal={Ain Shams Engineering Journal},
  volume={5},
  number={4},
  pages={1093--1113},
  year={2014}
}

@inproceedings{pang2002thumbs,
  title={Thumbs up? Sentiment Classification using Machine Learning Techniques},
  author={Pang, Bo and Lee, Lillian and Vaithyanathan, Shivakumar},
  booktitle={Proceedings of EMNLP},
  pages={79--86},
  year={2002}
}

@inproceedings{joachims1998text,
  title={Text Categorization with Support Vector Machines: Learning with Many Relevant Features},
  author={Joachims, Thorsten},
  booktitle={Proceedings of ECML},
  pages={137--142},
  year={1998}
}

@article{zhang2010understanding,
  title={Understanding bag-of-words model: a statistical framework},
  author={Zhang, Yin and Jin, Rong and Zhou, Zhi-Hua},
  journal={International Journal of Machine Learning and Cybernetics},
  volume={1},
  number={1},
  pages={43--52},
  year={2010}
}

@article{alzahrani2023survey,
  title={A survey of sentiment analysis: approaches, datasets, and open challenges},
  author={Alzahrani, Sawsan and others},
  journal={IEEE Access},
  volume={11},
  pages={101745--101780},
  year={2023}
}

@article{mikolov2013efficient,
  title={Efficient Estimation of Word Representations in Vector Space},
  author={Mikolov, Tomas and Chen, Kai and Corrado, Greg and Dean, Jeffrey},
  journal={arXiv preprint arXiv:1301.3781},
  year={2013}
}

@inproceedings{kim2014convolutional,
  title={Convolutional Neural Networks for Sentence Classification},
  author={Kim, Yoon},
  booktitle={Proceedings of EMNLP},
  pages={1746--1751},
  year={2014}
}

@article{hochreiter1997lstm,
  title={Long Short-Term Memory},
  author={Hochreiter, Sepp and Schmidhuber, J{"u}rgen},
  journal={Neural Computation},
  volume={9},
  number={8},
  pages={1735--1780},
  year={1997}
}

@inproceedings{devlin2019bert,
  title={BERT: Pre-training of Deep Bidirectional Transformers for Language Understanding},
  author={Devlin, Jacob and Chang, Ming-Wei and Lee, Kenton and Toutanova, Kristina},
  booktitle={Proceedings of NAACL-HLT},
  pages={4171--4186},
  year={2019}
}

@inproceedings{koto2020indonlu,
  title={IndoNLU: Benchmark and Resources for Evaluating Indonesian Natural Language Understanding},
  author={Koto, Fajri and Rahimi, Afshin and Lau, Jey Han and Baldwin, Timothy},
  booktitle={Proceedings of AACL-IJCNLP},
  pages={843--857},
  year={2020}
}

@book{goodfellow2016deep,
  title={Deep Learning},
  author={Goodfellow, Ian and Bengio, Yoshua and Courville, Aaron},
  year={2016},
  publisher={MIT Press}
}

@inproceedings{sculley2015hidden,
  title={Hidden Technical Debt in Machine Learning Systems},
  author={Sculley, D. and others},
  booktitle={Advances in Neural Information Processing Systems},
  volume={28},
  year={2015}
}

@article{lipton2018mythos,
  title={The Mythos of Model Interpretability},
  author={Lipton, Zachary C.},
  journal={Queue},
  volume={16},
  number={3},
  pages={31--57},
  year={2018}
}

\end{document}